\documentclass[letterpaper, 10 pt, journal, twoside]{ieeetran}
\usepackage{times}

\usepackage[numbers,sort]{natbib}
\usepackage{multicol}
\usepackage[bookmarks=true]{hyperref}

\usepackage{amsmath}
\usepackage{amssymb}
\usepackage{graphicx}
\usepackage{breqn}
\usepackage{subcaption}
\usepackage{xspace}
\usepackage{wrapfig}
\usepackage{relsize}
\usepackage{algorithm}
\usepackage{algorithmicx, algpseudocode}
\usepackage{multirow}

\usepackage{hyperref}
\usepackage[capitalise]{cleveref}

\usepackage{tabularx}
\usepackage{array}

\usepackage[hang,flushmargin]{footmisc}
\usepackage{url}

\usepackage[export]{adjustbox}
\usepackage{xcolor}         
\usepackage{float}

\newcommand{\Ours}{LEAGUE\xspace}

\newcommand{\shuo}[1]{\textcolor{green}{\xspace(SC: #1)}}

\usepackage{amsmath} 
\usepackage{amssymb}  

\usepackage{listings}

\usepackage{booktabs}
\usepackage{hyperref}
\usepackage{dblfloatfix}
\usepackage{graphicx} 
\usepackage{caption}
\usepackage{subcaption}

\newcommand{\kw}[1]{\textbf{#1}}

\usepackage{algorithm}
\usepackage[export]{adjustbox}
\algnewcommand\algorithmicdeclare{\textbf{Assume:}}
\algnewcommand\Declare{\item[\algorithmicdeclare]}

\usepackage{listings}
\usepackage{bm}

\begin{document}

\title{LEAGUE: Guided Skill Learning and Abstraction for Long-Horizon Manipulation}

\author{Shuo Cheng$^1$ and Danfei Xu$^1$
\thanks{Manuscript received: June, 23, 2023; Accepted August, 7, 2023.}
\thanks{This paper was recommended for publication by Jens Kober upon evaluation of the Associate Editor and Reviewers' comments.}
\thanks{$^1$Georgia Institute of Technology, correspondence: shuocheng@gatech.edu}
\thanks{Digital Object Identifier (DOI): see top of this page.}
}



%

\maketitle

\begin{abstract}
To assist with everyday human activities, robots must solve complex long-horizon tasks and generalize to new settings. Recent deep reinforcement learning (RL) methods show promise in fully autonomous learning, but they struggle to reach long-term goals in large environments. On the other hand, Task and Motion Planning (TAMP) approaches excel at solving and generalizing across long-horizon tasks, thanks to their powerful state and action abstractions. But they assume predefined skill sets, which limits their real-world applications. In this work, we combine the benefits of these two paradigms and propose an integrated task planning and skill learning framework named LEAGUE (\underline{Le}arning and \underline{A}bstraction with \underline{Gu}idanc\underline{e}). LEAGUE leverages the symbolic interface of a task planner to guide RL-based skill learning and creates abstract state space to enable skill reuse. More importantly, LEAGUE learns manipulation skills \emph{in-situ} of the task planning system, continuously growing its capability and the set of tasks that it can solve. We evaluate LEAGUE on four challenging simulated task domains and show that LEAGUE outperforms baselines by large margins. We also show that the learned skills can be reused to accelerate learning in new tasks domains and transfer to a physical robot platform.
\end{abstract}


\IEEEpeerreviewmaketitle

\section{Introduction}
\label{sec:intro}

\IEEEPARstart{D}{eveloping} robots that can autonomously learn to work in everyday human environments, such as households, has been a long-standing challenge.
Deep Reinforcement Learning (DRL) methods have shown promise in allowing robots to acquire skills with limited supervision~\cite{gu2017deep, levine2016end}, but they are still far from enabling home robots on their own.
Two significant challenges stand out: 1) real-world tasks are often long-horizon, requiring the learning agent to explore a vast space of possible action sequences that grows exponentially with the task duration, and 2) home robots must perform diverse tasks in varying environments, requiring the learner to either generalize or quickly adapt to new situations.


To better learn long-horizon tasks, many DRL methods propose to use domain knowledge and structural prior~\cite{bacon2017option, vezhnevets2017feudal, sharma2021autonomous}. Automatic goal generation in curriculum learning guides a learning process using intermediate subgoals, enabling an agent to explore and make incremental progress toward a long-horizon goal~\cite{sharma2021autonomous}. Other methods use skill primitives or learn hierarchical policies to enable temporally-extended decision-making ~\cite{nasiriany2022maple, bacon2017option}. Although these approaches can outperform vanilla DRL, they still suffer from low sample efficiency, lack of interpretability, and fragile generalization~\cite{bacon2017option, vezhnevets2017feudal}. Most importantly, the learned policies are often task-specific and fall short in cross-task and cross-domain generalization.


\begin{figure}
  \begin{center}
    \includegraphics[width=1\linewidth]{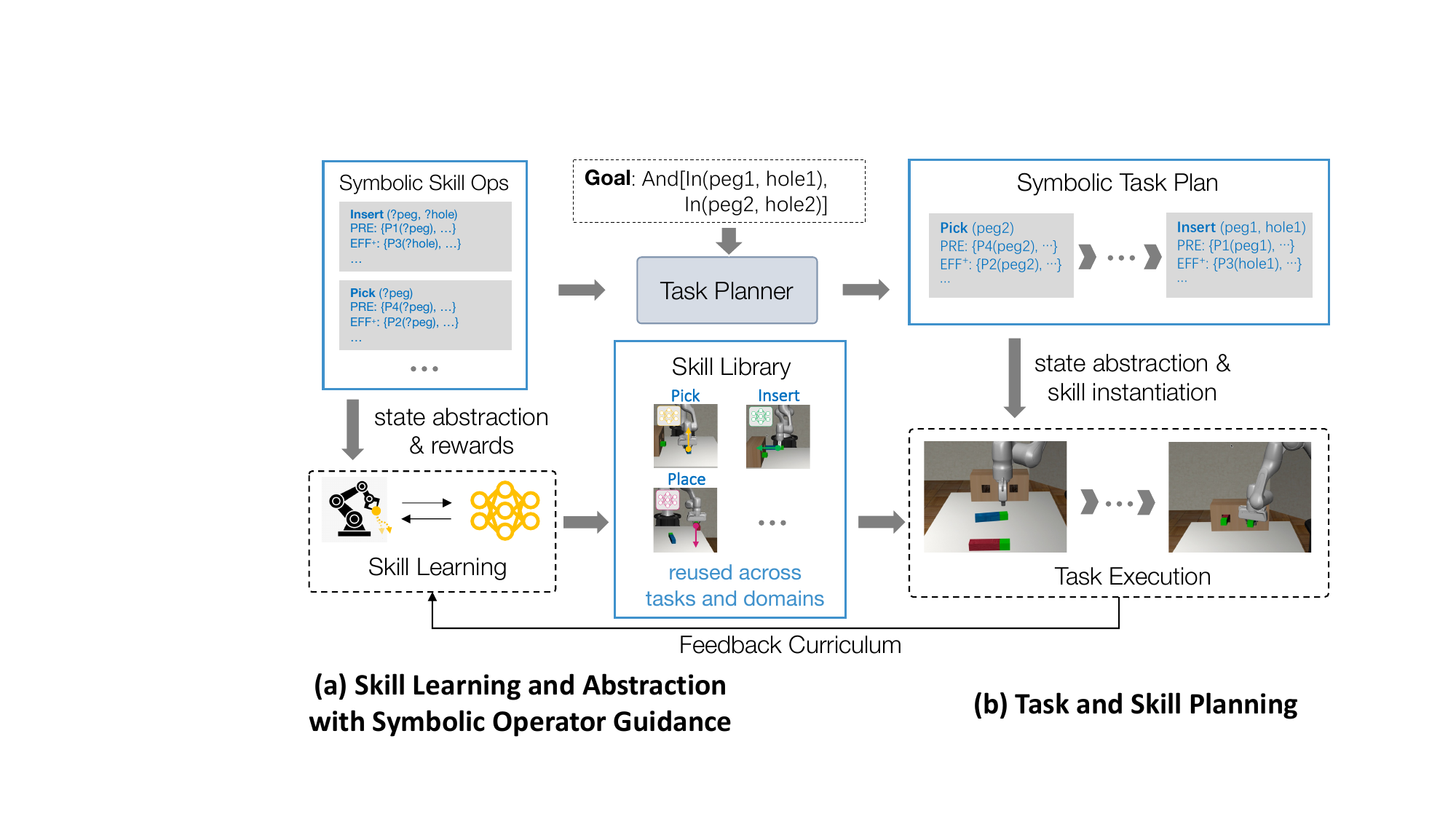}
    \vspace{-5pt}
  \end{center}
  \caption{\textbf{Overview of the \Ours framework.} We present an integrated task planning and skill learning framework. (a) The system uses the symbolic operator interface of a TAMP-like system as guidance to learn reusable manipulation skills (Alg. \ref{alg:sk_learn}). (b) A task planner composes the learned skills to solve long-horizon tasks (Alg.~\ref{alg:sk_plan}). As an integrated system, the task planner acts as a feedback curriculum (bottom) to guide skill learning, and the RL-based skill learner continuously grows the set of tasks that the system can solve.
    }
  \label{fig:overview}
  \vspace{-14pt}
\end{figure}

In the meantime, more established paradigms in robotics have long sought to address these challenges. In particular, Task and Motion Planning (TAMP)~\cite{kaelbling2011hierarchical,garrett2021integrated} leverages symbolic action abstractions to enable tractable planning and strong generalization. Specifically, the symbolic action operators divide a large planning problem into pieces that are each easier to solve. And the ``lifted'' action abstraction allows skill reuse across tasks and even domains. For example, a \texttt{grasp} skill operator and its underlying implementation can be easily adapted to solve a new task in a new domain. At the same time, most TAMP-style approaches assume access to a complete set of skills before deployment. This is impractical for two reasons. First, it is hard to prepare skills for all possible tasks. A robot must be able to grow its skill set on demand. Second, it is hard to hand-engineer manipulation skills for complex or contact-rich tasks (e.g., insertion). The challenges make TAMP methods difficult to deploy in real-world settings.

In this work, we introduce \Ours (\textbf{LE}arning and \textbf{A}bstraction with \textbf{GU}idanc\textbf{E}), an \emph{integrated task planning and skill learning} framework that learns to solve and generalize across long-horizon tasks (See Fig.~\ref{fig:overview}).
\Ours harnesses the merits of the two research paradigms discussed above. Starting with a task planner that is equipped with skills that are easy to implement (e.g., \texttt{reaching}), \Ours continuously grows the skill set \emph{in-situ} using a DRL-based learner. The intermediate goals in a task plan are prescribed as rewards for the learner to acquire and refine skills, and the mastered skills are used to reach the initial states of the new skills. Moreover, \Ours leverages the action operator definition, i.e., the preconditions and the effects, to determine a reduced state space for each learned skill, akin to the concept of information hiding in feudal learning~\cite{vezhnevets2017feudal}. The key idea is to \emph{abstract away} task-irrelevant features to make the learned skills modular and reusable. Together, the result is a \emph{virtuous cycle} where the task planner guides skill learning and abstraction, and the learner continuously expands the set of tasks that the system can solve.

We conduct empirical studies on four challenging long-horizon manipulation tasks built on the Robosuite simulation framework~\cite{robosuite2020}. We show that \Ours is able to outperform state-of-the-art hierarchical reinforcement learning methods~\cite{nasiriany2022maple} by a large margin. We also highlight that our method can achieve strong generalization to new task goals and even domains by reusing and adapting learned skills. As a result, \Ours can solve a challenging simulated coffee-making task where competitive baselines fall flat. We also demonstrate a LEAGUE system trained in simulation on a physical Franka Emika Panda robot. 

In summary, our primary contributions are: 1) we leverage the state and action abstractions readily available in a TAMP system to learn reusable skills, 2) we instantiate the synergies between the task planner and the skill learner as an integrated task planning and skill learning framework, and 3) we show that the framework can progressively learn skills to solve complex long-horizon tasks and generalize the learned skills to new task goals and domains.

\section{Related Work}
\label{sec:related}

\noindent\textbf{TAMP and Learning for TAMP.} 
Task and Motion Planning (TAMP)~\cite{kaelbling2011hierarchical,garrett2021integrated} is a powerful paradigm to solve long-horizon manipulation tasks. The key idea is to break a challenging planning problem into a set of symbolic-continuous search problems that are individually easier to solve. However, TAMP methods require high-level skills and their kinematics or dynamics models \emph{a priori}. The assumptions preclude domains for which hand-engineering manipulation skills is difficult, such as contact-rich tasks. Recent works proposed to learn dynamics models for TAMP by characterizing skill preconditions and effects~\cite{silverlearning,konidaris2018skills,liang2022search}. For example, Konidaris {et al.}~\cite{konidaris2018skills} learns compact symbolic models of an environment through trial-and-error. Liang {et al.}~\cite{liang2022search} uses graph neural networks to model skill effects. However, these works still require hand-engineering complete skill sets that can solve the target task, which may not be feasible in real-world applications. Our idea of learning skills to augment TAMP systems is closely related to Silver {et al.}~\cite{silverlearning}, which proposed to learn neural-symbolic skills via imitation. But they require access to hard-coded demonstration policies that can readily solve the target tasks. Our work instead aims to progressively grow TAMP skill libraries via guided reinforcement learning to solve long-horizon contact-rich manipulation tasks. 
 
\noindent\textbf{Curriculum for RL.}
Our idea to guide skill learning with a task planner is connected to curriculum-based RL, which is to expose an agent to incrementally more difficult intermediate tasks before mastering a target task~\cite{narvekar2020curriculum}. The intermediate tasks can take the form of environments~\cite{fang2020adaptive} and subgoals~\cite{sharma2021autonomous, uchendu2022jump}. For example, VaPRL~\cite{sharma2021autonomous} starts with near-success initialization and moves the initial states further away. While effective at accelerating task learning, existing curricula focus on teaching tasks or domain-specific policies. In contrast, our method leverages the symbolic abstraction of a task planner to learn a repertoire of modular and composable skills. We show that we can compose learned skills to achieve new goals and even transfer skills to new task domains. 

\noindent\textbf{State and Action Abstractions.}
State and action abstractions are crucial for learning tasks in a large environment~\cite{abel2020value}. State abstraction allows agents to focus on task-relevant features of the environment. Action abstraction enables temporally-extended decision-making for long-horizon tasks. There exists a large body of work on learning either or both types of abstractions~\cite{jonschkowski2015learning,konidaris2018skills,abel2020value,chitnis2021camps,xu2021deep,emmons2020sparse}. For example, Jonschkowski {et al.}~\cite{jonschkowski2015learning} explores different representation learning objectives for effective state abstraction. Abel {et al.}~\cite{abel2020value} introduces a theory for value-preserving state-action abstraction. However, autonomously discovering suitable abstractions remains an open challenge. Our key insight is that a TAMP framework provides powerful state and action abstractions that can readily guide skill learning. Specifically, the symbolic interface of an action operator defines both the precondition and the effect (action abstraction) and the state subspace that is relevant to the action (state abstraction). The abstractions allow us to train skills that are compatible with the task planner and prevent the learned skills from being distracted by irrelevant objects, making skill reuse across tasks and domains possible.

\noindent\textbf{Hierarchical Modeling in Robot Learning.}
Our method inherits the bi-level hierarchy of a TAMP framework. Hierarchical modeling has a rich history in robotics. In addition to TAMP, various general frameworks including hierarchical task networks~\cite{nejati2006learning,hayes2016autonomously,xu2018neural}, logical-geometric programming~\cite{toussaint2015logic}, and hierarchical reinforcement learning (HRL)~\cite{bacon2017option,vezhnevets2017feudal} have been proposed to exploit the hierarchical nature of common robotics tasks. In the context of HRL, a small number of works have explored symbolic planner-guided HRL~\cite{illanes2020symbolic}. However, these methods require tabular state representations and are thus limited to simple grid-world domains. In robotics domains, a closely related research thread is to use behavior primitives in RL~\cite{nasiriany2022maple,dalal2021accelerating}. For example, MAPLE~\cite{nasiriany2022maple} trains a high-level policy that chooses hand-engineered behavior primitives and atomic actions. Our method instead leverages a symbolic planner to serve as the high-level controller to compose learned skills, allowing us to continuously extend the skill set while also leading to better generalization.
\section{Method}
\label{sec:method}

We seek to enable robots to solve and generalize across long-horizon tasks. Our primary contribution is a novel integrated task planning and skill learning framework named \Ours. Here, we first provide the necessary background in Sec.~\ref{ssec:background}, and describe how \Ours (1) learns reusable skills guided by the symbolic operators of a task planner in Sec.~\ref{ssec:skill_learning} and (2) uses planner-generated task plans as an autonomous curriculum to continuously learn skills and expand the capability of the overall system in Sec.~\ref{ssec:system}.  

\vspace{-5pt}
\subsection{Background}
\label{ssec:background}
\noindent\textbf{MDP.} We consider a Markov Decision Process (MDP) $<\mathcal{X}, \mathcal{A}, \mathcal{R}(x, a), \mathcal{T}(x'|x, a), p(x^{(0)}), \gamma>$, with continuous state space $\mathcal{X}$, continuous action space $\mathcal{A}$, reward function $\mathcal{R}$, and environment transition model $\mathcal{T}$. $p(x^{(0)})$ denotes the distribution of the initial states, $x^{(H)}$ denotes terminal state, and $\gamma$ is the discount factor. The objective for RL training is to maximize the expected total reward of the policy $\pi(a|x)$ that the agent uses to interact with the environment: $J = \mathbb{E}_{x^{(0)}, a^{(0)}, ..., x^{(H)} \sim\pi, p(x^{(0)})} \left[ \sum_t \gamma^t \mathcal{R}(x^{(t)}, a^{(t)}) \right] \label{eq:rl}$.

\noindent\textbf{Task planning space.} To support task planning, we assume the environment is augmented with a symbolic interface $<\mathcal{O}, \Lambda, \Bar\Psi, \Bar\Omega, \mathcal{G}>$, where $\mathcal{O}$ denotes the object set and $\Lambda$ denotes a finite set of object types. Each object entity $o \in \mathcal{O}$ (e.g., \texttt{peg1}) has a specific type $\lambda \in \Lambda$ (e.g., \texttt{peg}) and a tuple of $\texttt{dim}(\lambda)$-dimensional feature containing information such as poses and joint angles, and the environment state $x \in \mathcal{X}$ is a mapping from object entities to features: $x(o) \in \mathbb{R}^{\texttt{dim}(\texttt{type}(o))}$. Predicates $\Bar\Psi$ describe the relationships among multiple objects. Each predicate $\Bar\psi$ (e.g., \texttt{Holding(?object:peg)}) is characterized by a tuple of object types $(\lambda_1, ..., \lambda_m)$ and a binary classifier that determines whether the relationship holds: $c_{\Bar\psi}: \mathcal{X}\times\mathcal{O}^{m} \rightarrow \{True, False\}$, where each substitute entity $o_i \in \mathcal{O}$ is restricted to have type $\lambda_i \in \Lambda$. Evaluating a predicate on the state by substituting corresponding object entities will result in a ground atom (e.g., \texttt{Holding(peg1)}). A task goal $g \in \mathcal{G}$ is represented as a set of ground atoms, where a symbolic state $x_{\Psi}$ can be obtained by evaluating a set of predicates $\Bar\Psi$ and keeping all positive ground atoms: 

\begin{multline}
\begin{aligned}
x_{\Psi} &= \textsc{parse}(x, \mathcal{O}, \Bar\Psi) \\
&\overset{\triangle}{=} \{\underline{\psi}: c_{{\Bar\psi}}(x, \mathcal{O}^{\Bar\psi}) = True, \forall \mathcal{O}^{\Bar\psi} \subseteq \mathcal{O}, \forall \Bar\psi \in \Bar\Psi\} 
\end{aligned}
\label{eq:parse}
\end{multline}

where $\mathcal{O}^{\Bar{\psi}}$ is a subset of object entities that each entity $o_i$ has the same object type $\lambda_i$ specified by the predicate $\Bar\psi$.

\noindent\textbf{Symbolic skill operators.} Following prior works~\cite{garrett2021integrated}, we characterize lifted skill operator $\Bar{\omega} \in \Bar{\Omega}$ by a tuple $<\textsc{PAR}, \textsc{PRE}, \textsc{EFF}^{+}, \textsc{EFF}^{-}>$, where $\textsc{PRE}$ denotes the precondition of the operator, which is a set of lifted atoms defining the condition that the operator is executable. $\textsc{EFF}^{+}$ and $\textsc{EFF}^{-}$ are lifted atoms that describe the expected effects (changes in conditions) upon successful skill execution. $\textsc{PAR}$ is an ordered parameter list that defines all object types used in $\textsc{PRE}$, $\textsc{EFF}^{+}$, and $\textsc{EFF}^{-}$. A ground skill operator $\underline{\omega}$ substitutes lifted atoms with object instances: $\underline{\omega} = ⟨\Bar{\omega}, \delta⟩ \overset{\triangle}{=} <\underline{\textsc{PRE}}, \underline{\textsc{EFF}^{+}}, \underline{\textsc{EFF}^{-}}>$, where $\delta: \Lambda \rightarrow \mathcal{O}$. Given a task goal, a symbolic task plan is a list of ground operators that, when the instantiated skills are executed successfully, lead to an environment state that satisfies the goal condition.

As a running example, consider a short task of inserting a peg (\texttt{peg1}) into the target hole (\texttt{hole1}). The applicable operators for this task are defined as:



\begin{lstlisting}
Pick|(?object)|
  |\kw{PAR}:| [?object:peg]
  |\kw{PRE}:| {HandEmpty(),OnTable(?object)}
  |\kw{EFF$^-$}:| {HandEmpty(),OnTable(?object)}
  |\kw{EFF$^+$}:| {Holding(?object)}
Insert|(?object,?hole)|
  |\kw{PAR}:| [?object:peg,?hole:hole]
  |\kw{PRE}:| {Holding(?object),IsClear(?hole)}
  |\kw{EFF$^-$}:| {Holding(?object),IsClear(?hole)}
  |\kw{EFF$^+$}:| {HandEmpty(),In(?object,?hole)}

\end{lstlisting}



The environment starts with \texttt{peg1} on the table.
Evaluating the \textsc{PARSE} function (Eq.~\ref{eq:parse}) 
yields a symbolic state $\{\texttt{HandEmpty(),}\texttt{IsClear(hole1),\texttt{OnTable(peg1)}}\}$, a set of grounded atoms that satisfies the preconditions of the grounded operator \texttt{Pick(peg1)}. This grounded operator, if executed successfully, should reach the symbolic state of $\{\texttt{Holding(peg1),}\texttt{IsClear(hole1)}\}$, which is an intermediate subgoal for the final task goal that is characterized by the grounded atoms $\{\texttt{HandEmpty(),}\texttt{In(peg1,hole1)}\}$. The symbolic task plan is therefore $\mathcal{P} =[\texttt{Pick(peg1),Insert(peg1,hole1)}]$.

We are interested in learning primitive manipulation skills
for accomplishing individual subgoals induced by the expected effects of the corresponding operators -- the building blocks that
constitute a symbolic task plan. In our setting, each lifted operator $\bar{\omega}$ will have a corresponding skill policy $\pi$ to be learned, while during execution the ground operators belong to the same lifted operator $\bar{\omega}$ share the same skill policy. We assume access to the predicates $\Bar\Psi$ and the lifted operators $\Bar\Omega$ of the environments and focus on efficiently learning the skills for achieving the effects. Note that it is possible to invent and learn predicates and operators~\cite{silver2021learning,silver2023inventing}, but the topics are beyond the scope of this work.



\vspace{-3pt}
\subsection{Skill Learning and Abstraction with Operator Guidance}
\label{ssec:skill_learning}
Action and state abstractions are fundamental to TAMP's abilities to solve and generalize across long-horizon tasks~\cite{garrett2021integrated}. Our key insight is that these abstractions, in the form of symbolic action operators, can readily guide RL-trained policies to gain similar abilities. Specifically, for action abstraction, we train temporally-extended skills to reach desired effects of a skill operator by prescribing the effect condition as shaped reward. For state abstraction, we take inspiration from the idea of \emph{information hiding} in feudal learning~\cite{vezhnevets2017feudal} and use the precondition and effect signature of an operator to determine a \emph{skill-relevant} state space for its corresponding learned policy. This allows the policy to be robust against domains shift and achieve generalization, especially in large environments where most elements are impertinent to a given skill. To further accelerate skill learning, we leverage the existing motion planning capability of a TAMP system to augment the learned skill with a transition primitive. Below we describe each component in detail. 


\noindent\textbf{Symbolic operators as reward guidance.}
\label{ssec:reward_guidance}
Our skill learner leverages the existing RL method that supports continuous action space. In this work, we use Soft Actor-Critic (SAC)~\cite{haarnoja2018soft} as the basis for skill learning. SAC leverages entropy regularization to enhance exploration. Given the ground operator $\underline{\omega}$ of a skill, we can define an operator-guided reward $\mathcal{R}_\Psi$ for each individual skill based on continuous environment state ${x}$ and the action $a$ produced by the corresponding policy $\pi$ that takes in skill-related state $\hat{x}$ (which will be described later), the objective for our skill learning is therefore rewritten as: 

\begin{multline}
J = \mathbb{E}_{{x}^{(0)}, a^{(0)}, ..., {x}^{(H)}  \sim\pi, p(x^{(0)})}
\left[\begin{aligned} 
\sum_t \gamma^t (\mathcal{R}_\Psi({x}^{(t)}, a^{(t)}, \underline{\omega}) \\ +\alpha\mathcal{H}(\pi(\cdot|\hat{x}^{(t)}))
\end{aligned}\right] 
\label{eq:our_rl}
\end{multline}

where $R_\Psi(\cdot) \mapsto [0, 1]$, $\mathcal{H}(\cdot)$ is the entropy term introduced by SAC. While it is possible to learn directly from sparse reward defined by the symbolic state, in practice we associate each operator-guided reward with a dense reward function implemented in the Robosuite~\cite{robosuite2020} benchmark for better learning efficiency. Continuing our running example, the shared reward for \texttt{Pick} is defined as $1 - \tanh(10.0 * d)$, where $d$ is the distance between the gripper center and target object center, and the target object is identified by the task planner. 

\noindent\textbf{Enhance skill reuse with feudal state abstraction.} 
\label{ssec:state_abs}
With the precondition and effect signature of a ground operator $\underline{\omega}$, we can determine a skill-relevant state space to further prevent the learned policy from being distracted by task-irrelevant objects: 
\begin{equation}
\hat{x} = \textsc{extract}(x, \underline{\omega}, \mathcal{O}) \overset{\triangle}{=} \{x(o): o \in \underline{\textsc{PAR}}, \forall o \in \mathcal{O}\}
\end{equation}
 where $\underline{\textsc{PAR}}$ is the parameter list of the ground operator. In our running example, the skill-related state $\hat{x}$ for \texttt{Pick(peg1)} includes the 6D pose of \texttt{peg1} and the state of the robot. This design echoes previous works that learn to impose constraints on states~\cite{chitnis2021camps}, except that here the constraints are directly informed by the task planner.

\noindent\textbf{Accelerate learning with transition motion primitives.} A key to our method is learning modular manipulation skills that can be composed to solve long tasks. However, for complex manipulation problems, even learning such short skills can be challenging. On the other hand, although TAMP systems fall short when facing contact-rich manipulation, they excel at finding collision-free paths. To this end, we propose to augment our policy with motion planner-based transition primitives. The key idea is to first approach the skill-relevant object (per the skill operator) using an off-the-shelf motion planner, before convening RL-based skill learning. For the target of motion planning, we simply set the goal position to be 0.04m higher than the object or placement position of interest that was identified by the task planning. The component significantly speeds up the exploration while still allowing the system to learn closed-loop contact-rich manipulation skills. 

\vspace{-5pt}
\subsection{Integrated Task Planning and Skill Learning}
\label{ssec:system}

So far, we have described a recipe for learning reusable skills using symbolic skill operators as guidance. But these skills are not learned in silos. A key to \Ours's success is to learn skills \emph{in-situ} of a task planning system. The integrated planning and learning scheme ensures that the learned skills are compatible with the planner, and the skill learner can continuously extend the capability of the overall system to solve more tasks. Here we first describe how \Ours performs task planning and execution at inference time, and then we introduce an algorithm that uses task plans as an autonomous curriculum to schedule skill learning. 

\noindent\textbf{Task planning and skill execution.}
To plan for task goal $g$, we first \textsc{parse} (Eq.~\ref{eq:parse}) the continuous environment state $x$ for obtaining symbolic state $x_{\Psi}$, which affords symbolic search with ground operators. We then {ground} each lifted operator $\Bar\omega \in \Bar{\Omega}$ on the object set $\mathcal{O}$ by substituting object entities in preconditions and effects, leading to ground operators $\underline{\omega} = <\underline{\textsc{PRE}}, \underline{\textsc{EFF}^{+}}, \underline{\textsc{EFF}^{-}}>$ that support operating with symbolic states. A ground operator is considered executable only when its preconditions are satisfied: $ \underline{\textsc{PRE}} \subseteq x_{\Psi} $. The operators induce an abstract transition model $F(x_{\Psi}, \underline{\omega})$ that allows planning in symbolic space: 
\begin{equation}
x_{\Psi}' = F(x_{\Psi}, \underline{\omega}) \overset{\triangle}{=} (x_{\Psi} \setminus \underline{\textsc{EFF}^{-}}) \cup \underline{\textsc{EFF}^{+}}    
\end{equation}
We use PDDL~\cite{fox2003pddl2} to build the symbolic planner and use A$\ast$ search for generating high-level plans. 

With the generated task plan, we sequentially invoke the corresponding skill $\pi^*$ to reach the subgoal that complies with the effects of each ground operator $\underline{\omega}$ in the plan. We {rollout} each skill controller until it fulfills the effects of the operator or a maximum skill horizon $H$ is reached. To {verify} whether the $l$-th skill is executed successfully, we first obtain the corresponding symbolic state $x_{\Psi}^{l}$ by parsing the ending environment state $x^{*}$. The execution is considered successful only when the environment state $x^{*}$ conforms to the expected effects: $F(x_{\Psi}^{l-1}, \underline{\omega}_l) \subseteq x_{\Psi}^{l}$. We keep track of the failed skills and the starting simulator info $s^*$ to inform the learning curriculum.

\noindent\textbf{Task planner as an automated curriculum.}
To efficiently acquire all necessary skills for a given multi-step task, we leverage the task planner as an automated curriculum to learn skills in a progressive manner. The key idea is to use more proficient skills to reach the preconditions of skills that require additional learning (See Fig.~\ref{fig:overview}). 
The algorithm is sketched in Alg.~\ref{alg:sk_learn} and Alg.~\ref{alg:sk_plan}. On a high level, we repeat task planning and skill learning until convergence. We keep track of failed skills during $N$ task executions and adopt strict scheduling criteria, where a skill is scheduled for learning (Sec.~\ref{ssec:skill_learning}) if it ever fails during the $N$ episodes.  
Notably, we share the replay buffers for different skill instances (e.g., \texttt{Pick(peg1)} and \texttt{Pick(peg2)}) that belong to the same lifted operator,  so that the relevant experience can be reused to further improve the learning efficiency and generalization. 
\begin{algorithm}
\scriptsize
\caption{\textsc{SkillCurriculum}}\label{alg:sk_learn}
\begin{algorithmic}
\State \textbf{hyperparameters:} \\
Number of training iterations $K$
\State \textbf{input:} \\
\texttt{env}  \Comment{task environment} \\
$g$  \Comment{symbolic task goal} \\
$\Bar{\Psi}$ \Comment{state predicates} \\
$\Bar\Omega$ \Comment{lifted operators}
\State \textbf{start} \\
$\Pi \leftarrow [\pi_{1}^{(0)}, ..., \pi_{|\Bar\Omega|}^{(0)}]$ \Comment{initialize all skill policies} \\
$t \leftarrow 0$
\While {\emph{Not Converged}}
\State $\mathcal{D}\leftarrow \emptyset$ \Comment{buffer for failed skills}
\For{$i \leftarrow [1, ..., N]$}
\State $\mathcal{D} \leftarrow \mathcal{D} \cup
\textsc{TrySolveTask}(\texttt{env}, g, \Bar{\Psi}, \Bar\Omega, \Pi)$
\EndFor
\For{$i, s, \underline{\omega} \leftarrow \mathcal{D}$}
\State $\pi_i^{(t)} \leftarrow \Pi[i]$
\For{$k \leftarrow [1, ..., K]$}
\State $\pi_{i}^{(t+k)} \leftarrow\textsc{SAC}(\texttt{env}, s, \pi_{i}^{(t+k-1)}, \underline{\omega})$ 
\Comment{RL training}
\EndFor
\State $\Pi[i] \leftarrow \pi_{i}^{(t+K)}$
\EndFor
\State $t \leftarrow t+K$
\EndWhile
\State \Return $\Pi$
\end{algorithmic}
\end{algorithm}

\begin{algorithm}
\scriptsize
\caption{\textsc{TrySolveTask}}\label{alg:sk_plan}
\begin{algorithmic}
\State \textbf{hyperparameters:} \\
Maximal skill horizon $H$ 
\State \textbf{input:} \\
\texttt{env}  \Comment{task environment} \\
$g$  \Comment{symbolic task goal} \\
$\Bar\Psi$ \Comment{state predicates} \\
$\Bar\Omega$ \Comment{lifted operators} \\
$\Pi$ \Comment{skill policies}
\State \textbf{start}\\
$\mathcal{O}, x^{(0)} \leftarrow\texttt{env.get_state()}$ \\
$x_\Psi^{(0)} \leftarrow\textsc{parse}(x^{(0)}, \mathcal{O}, \Bar\Psi)$ \Comment{continuous state to symbolic state}\\
$\underline{\Omega} \leftarrow \textsc{ground}(\mathcal{O}, \Bar\Omega)$ \Comment{get grounded operators}\\
$[\underline{\omega}_1, ..., \underline{\omega}_L] \leftarrow \textsc{search}(x_\Psi^{(0)}, g, \underline{\Omega})$ \Comment{found plan with length L} \\
$\mathcal{D}, l \leftarrow [], 0$\\
\While{$l < L$}
\State $i \leftarrow \textsc{LookUpSkill}(\underline{\omega}_l)$
\State $\pi^* \leftarrow \Pi[i]$
\State $s^* \leftarrow \texttt{env.get_sim()}$ \Comment{get simulator state}
\State $x^* \leftarrow \textsc{rollout}(\texttt{env}, \pi^*, H)$

\If{$\textsc{IsSuccess}(x^*, \underline{\omega}_l)$}
    \State $l \leftarrow l + 1$ \Comment{advance to the next skill}
    \State $\textbf{continue}$
\Else{}
    \State $\mathcal{D} \leftarrow \mathcal{D}\cup(i, s^*, \underline{\omega}_l)$ \Comment{collect failed skills and states}
\EndIf

\EndWhile
\State \Return $\mathcal{D}$
\end{algorithmic}
\end{algorithm}



\begin{figure*}
    \centering
    \includegraphics[width=0.9\linewidth]{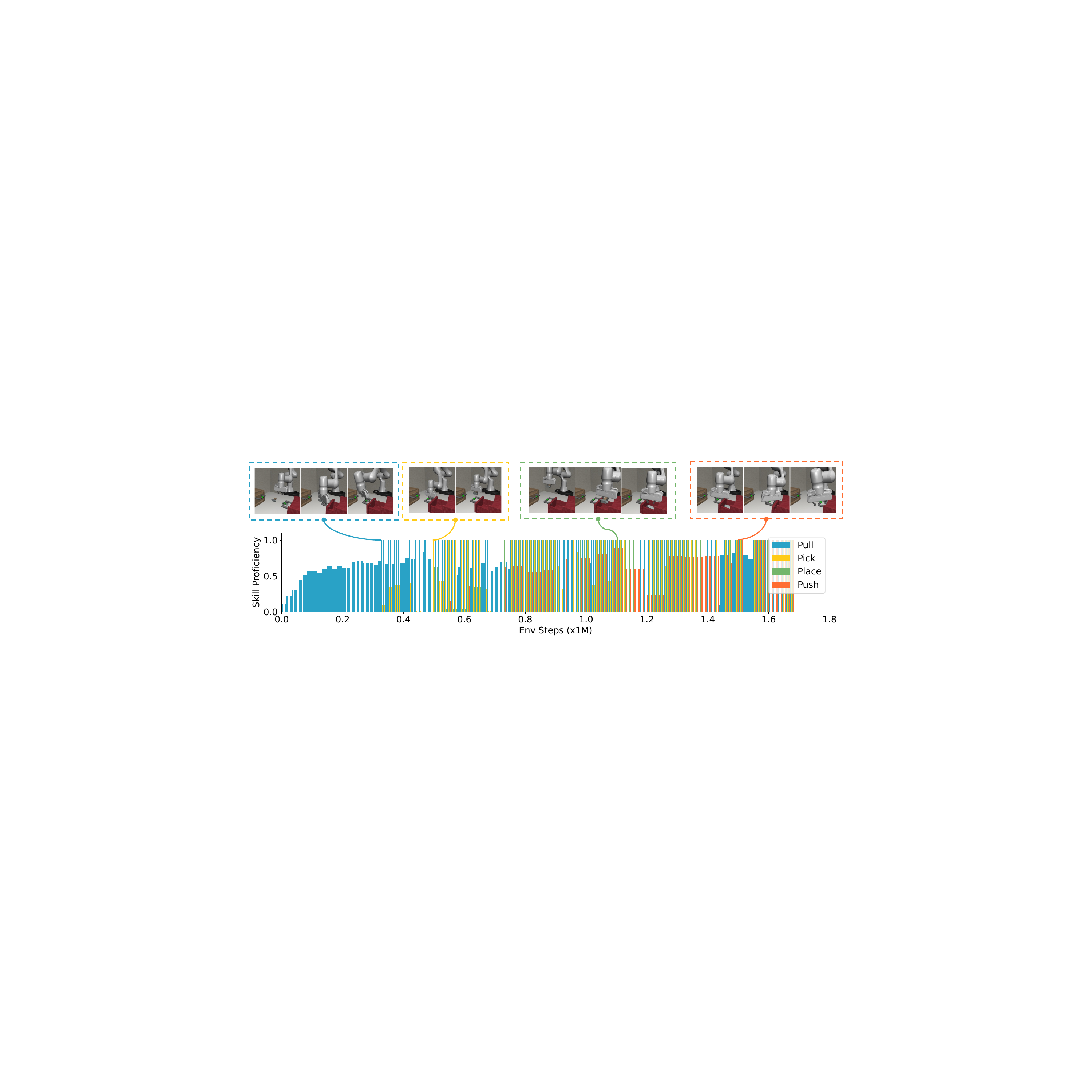}
    \caption{\textbf{Visualizing skill learning progress.} The plot shows the proficiency level of each skill throughout the process of learning a simplified {StowHammer} task. The skill proficiency is the average normalized reward a skill receives at an iteration.}
    \label{fig:skill_switch}
    \vspace{-8pt}
\end{figure*}

\vspace{-5pt}
\section{Experiments}
\label{sec:exp}
Our experiments aim to show that 1) \Ours can progressively learn and refine skills to solve long-horizon tasks and 2) our novel operator-guided skill learning and abstraction algorithm produces composable and reusable skills, enabling quick adaptation to new tasks and domains. Finally, we demonstrate transferring a trained \Ours system to a physical robot.


\subsection{Experimental Setup}
\begin{figure}
  \begin{center}
    \includegraphics[width=0.8\linewidth]{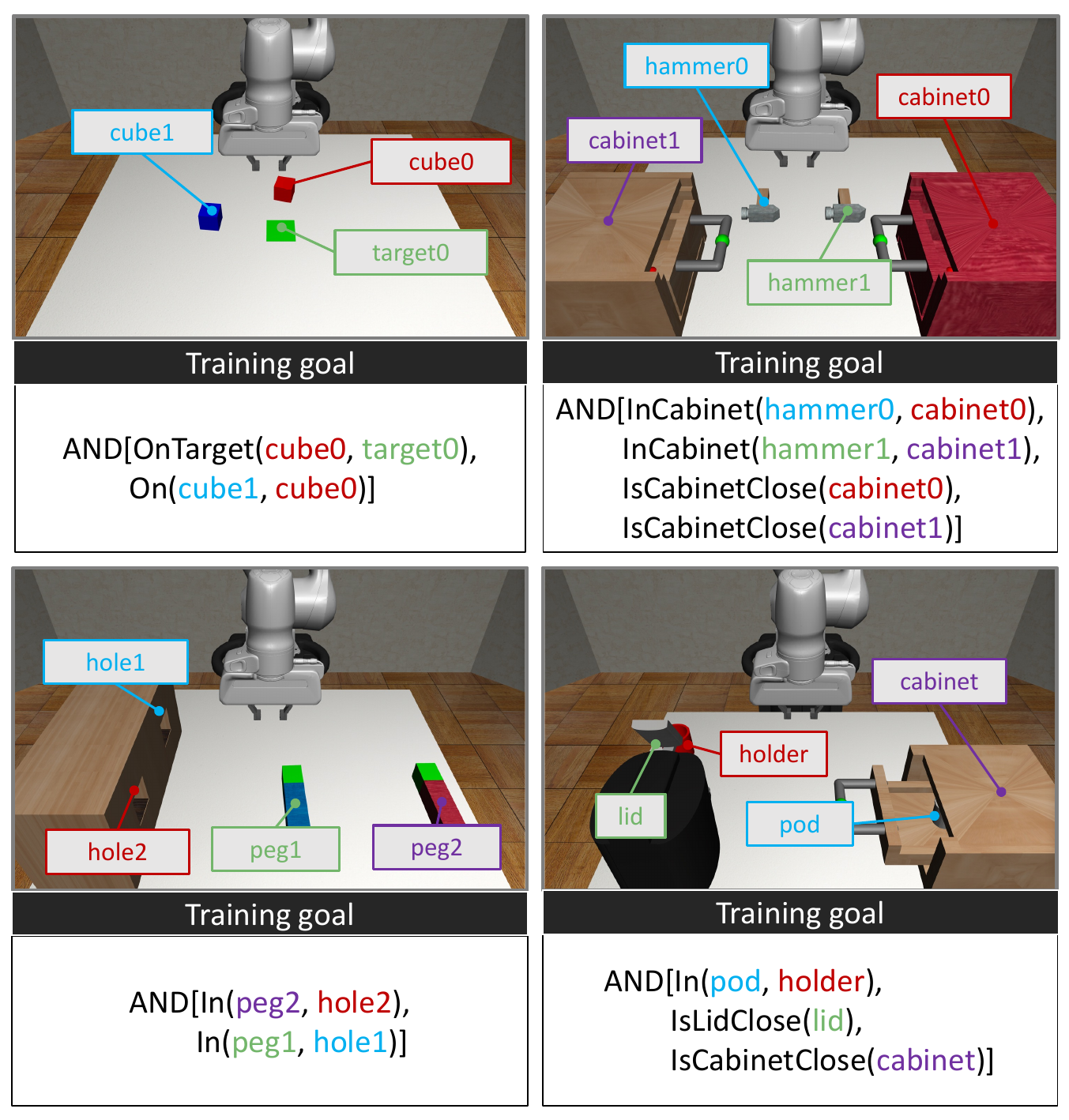}
    \vspace{-5pt}
  \end{center}
  \caption{\textbf{Simulation task setup.} We show the initialization of the simulation setups for the four tasks: {StackAtTarget}, {StowHammer}, {PegInHole} and {MakeCoffee}.}
  \label{fig:sim_setup}
      \vspace{-20pt}
\end{figure}

We conduct evaluations in four simulated domains, in which we devise tasks that require multi-step reasoning, contact-rich manipulation, and long-horizon interactions (See Fig.~\ref{fig:sim_setup}).

\textbf{StackAtTarget} is to stack two cubes on a tight target region with a specific order. The applicable skill operators are \texttt{Pick} and \texttt{Place}.  Since the cubes are randomly placed in the scene, the top cube may occupy the target region, in which situation the robot must first remove the top cube before stacking. 

\textbf{StowHammer} requires the robot to stow two hammers into different closed cabinets. It involves four skills: \texttt{Pick}, \texttt{Place}, \texttt{Pull}, \texttt{Push}. Since the workspace is tight, the robot needs to close an opened cabinet before being able to open the other one, which requires multi-step reasoning. 

\textbf{PegInHole} is to pick up and insert two pegs into two horizontal holes. The applicable operators are \texttt{Pick} and \texttt{Insert}. This task challenges the robot with contact-rich manipulations and multi-step planning. 

\textbf{MakeCoffee} is to pick up a coffee pod from a closed cabinet, insert it into the holder of the coffee machine, and finally close both the lid and the cabinet, The applicable operators are \texttt{Pick}, \texttt{Pull}, \texttt{Push}, \texttt{CloseLid}, and \texttt{InsertHolder}. 

The environments are built on Robosuite~\cite{robosuite2020} simulator. We use a Franka Emika
Panda robot arm that is controlled at 20Hz with an operational space controller (OSC), which has 5 degrees of freedom: end-effector position and the yaw angle and the position of the gripper. See Fig.~\ref{fig:sim_setup} for an illustration. 

\begin{figure*}
  \centering
  \includegraphics[width=0.8\linewidth]{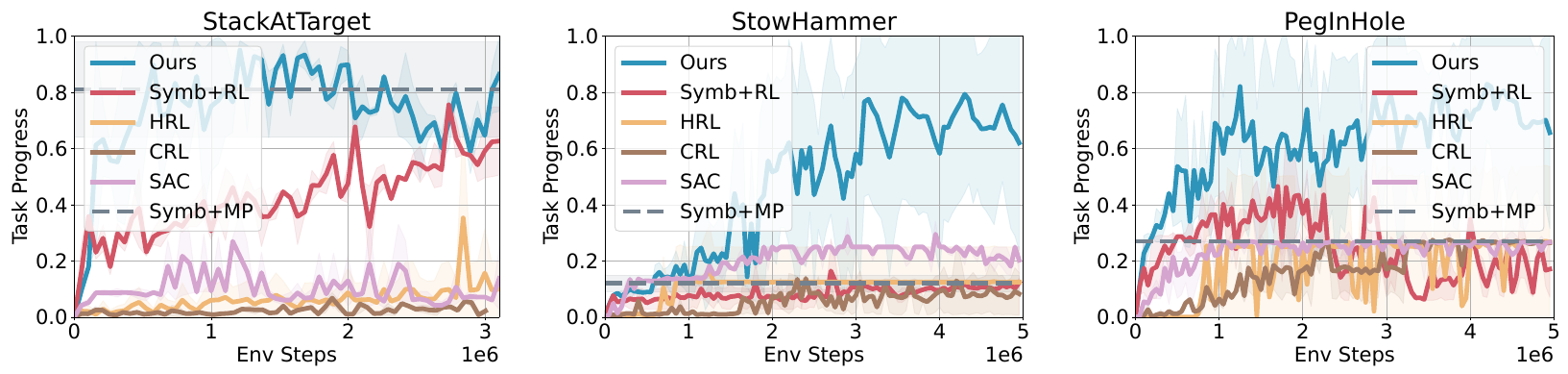}
  \caption{\textbf{Baseline comparison.} We compare relevant methods on three task domains. The plot shows the corresponding average task progress during evaluation throughout training, which is measured as the summation of achieved rewards of each successfully executed skill in the task plan and normalized to 1. The results are reported using 5 random seeds, with the standard deviation shown as the shaded area.}
  \label{fig:main_exp}
    \vspace{-5pt}
\end{figure*}


\subsection{Visualize the Progressive Skill Learning Process}
Before discussing quantitative comparisons, we seek to gain intuitive understanding of our progressive skill learning scheme (Sec.~\ref{ssec:system}), where the learning curriculum adjusts based on the proficiencies of the skills. In Fig.~\ref{fig:skill_switch}, we visualize the proficiency level of each skill throughout the process of learning a simplified {StowHammer} task, where the goal is to stow away one hammer instead of two. The $y$ axis is the average normalized reward a skill receives. Note that we only visualize a subset of skills scheduled for training at an iteration. 
The corresponding behavior of each skill at a certain stage is visualized in the snapshots on top of the plot. 

At the beginning of the training, the system can only reach the precondition for executing the \texttt{Pull(?cabinet)} skill but not other skills, thus the experience of \texttt{Pull(?cabinet)} skill is collected and it is repeatedly selected for training. Until the agent is able to open one of the cabinets, the second planned skill \texttt{Pick(?object)} is then instantiated for learning and execution. Finally at the end of the training, all skills become proficient to be used to execute the entire task. The result qualitatively shows that \Ours's automated curriculum is effective at progressively learning skills to achieve long-horizon task goals. 

\subsection{Quantitative Evaluation}

Here, we seek to highlight various aspects of our solution paradigm through quantitatively comparing \Ours with a number of strong baseline methods. Below we describe the baselines and discuss the results.

\begin{itemize}
    \item \textbf{RL (SAC)}: We adopt SAC~\cite{haarnoja2018soft} as a strong RL baseline. To facilitate a fair comparison, we extend the vanilla task reward function to staged rewards using an oracle task plan, where the reward at each step is the summation of achieved rewards for each completed subgoal and the reward for the current subgoal.
    \item \textbf{Curriculum RL (CRL)}: We follow the main idea of state-of-the-art curriculum RL approaches~\cite{sharma2021autonomous, uchendu2022jump}, which starts the training with near-success initializations and gradually move the initial states back to the true environment initial states. To facilitate a fair comparison, we sample the curriculum's initial states based on the subgoals of an oracle task plan (in reverse) and adopt the same staged reward described above.
    \item \textbf{Hierarchical RL (HRL)}: This baseline adapts the recent primitive-based HRL frameworks~\cite{nasiriany2022maple, dalal2021accelerating} for our tasks. The key idea is to train a high-level meta controller to compose parameterized skill primitives and atomic actions. We base our implementation on MAPLE~\cite{nasiriany2022maple} and use the oracle task plan to identify the target objects for defining the affordance to guide the exploration. 
    \item \textbf{Symb+MP}: An open-loop baseline that resembles a vanilla TAMP framework, which greedily generates a motion plan for each skill in a task plan. The robot then executes the plan through a trajectory controller.  
    \item \textbf{Symb+RL}: An ablation baseline of \Ours that removes the state abstraction (~\ref{ssec:state_abs}) and retains all other features including the symbolic plan-based curriculum. 
\end{itemize}


The multi-stage nature of our evaluation tasks makes designing smooth task-level metrics difficult. Thus we adopt \textbf{task progress} as our metric, which is defined as the summed reward of all task stages normalized to $[0,1]$. Below we discuss the main findings based on Fig.~\ref{fig:main_exp}. 

\textbf{High-level reasoning is critical for solving long-horizon tasks.} We observe that in {StackAtTarget}, a long-horizon task with relatively simple manipulation steps, methods equipped with a task planner ({\Ours}, {Symb+MP}, and {Symb+RL}) significantly outperforms all other baselines. The most competitive {HRL} baseline occasionally learns to move the bottom cube to the target region. This shows the value of explicit high-level reasoning, in particular as a plan-informed automated curriculum in \Ours. Notably, the open loop {Symb+MP} performs on par with \Ours because simple picking and placing can readily be solved by open-loop trajectories.

\textbf{\Ours can solve long-horizon, contact-rich manipulation tasks.} \Ours significantly outperforms all other baselines in {StowHammer} and {PegInHole}, which are both long-horizon and require contact-rich manipulation. Notably, most baselines cannot advance beyond opening the cabinet in {StowHammer} and picking up the first peg in {PegInHole}.



\textbf{Skill reuse is critical to learning structured tasks.} Common multi-step tasks have repeating structures, which can be leveraged by methods that explicitly reuse learned skills. We note that both \Ours and Symb+RL perform competitively in {StackAtTarget} that involve repeating steps (i.e., stack two cubes). On the other hand, {HammerPlace} and {PegInHole} involve more objects, most of which are not relevant to a given skill. This prevents na\"ive skill reuse --- a policy may learn spurious correlation to these irrelevant features --- and necessitates state abstraction, which we will discuss next.

\textbf{State abstraction facilitates skill reuse in complex environments.} We observe that \Ours outperforms Symb+RL in both {HammerPlace} and {PegInHole}. This shows that state abstraction can further improve skill reuse in complex environments by ignoring features that are irrelevant to a skill. We will also show in Sec.~\ref{sec:new task and domain} that skill reuse enables our method to generalize to novel task goals and domains.

\textbf{Other observations.} We observe that without explicit prior structures such as motion primitive, {SAC} baseline is able to exploit environment artifacts and learn shortcut behaviors. For example, in the {StowHammer} task, SAC agent learns to grip the head of the hammer to prevent slipping, but the grasping pose precludes it from fitting the hammer to the drawer. 
Moreover, our analysis found that the CRL agent often failed to reach the final goal from some intermediate states due to the strong sequential dependency of our evaluation tasks: the robot must succeed in one stage to reach the pre-condition of the next. And because the environment steps budget is distributed to multiple stages, CRL often underperforms other baselines (e.g., SAC) in completing the initial stages of a task. 

\subsection{Generalization to New Tasks and Domains} \label{sec:new task and domain}
\begin{figure}
\vspace{-5pt}
\makeatletter\def\@captype{table}\makeatother
\caption{We report the performance of applying our method to new task goals in the {StowHammer} and the {PegInHole} domains without additional learning.}
\resizebox{1\linewidth}{!} {
    \begin{tabular}{|c|c|c|c|}
    \hline
    \multicolumn{1}{|l|}{} & Training Goal      & Test Goal1      & Test Goal2    \\ \hline
\textbf{StowHammer}                  & 0.94 $\pm$ 0.21 & 0.90 $\pm$ 0.12 & 0.73 $\pm$ 0.31  \\ \hline
\textbf{PegInHole}                & 0.87 $\pm$ 0.23 & 0.53 $\pm$ 0.05 & 1.00 $\pm$ 0.00  \\ \hline
    \end{tabular}
  }
\label{tab:new_task}
\vspace{-5pt}
\end{figure}

To validate that our method can effectively generalize to new task goals and even new task domains by reusing learned skills, we present the following experiments.
\begin{figure*}[t]
  \centering
  \includegraphics[width=0.95\linewidth]{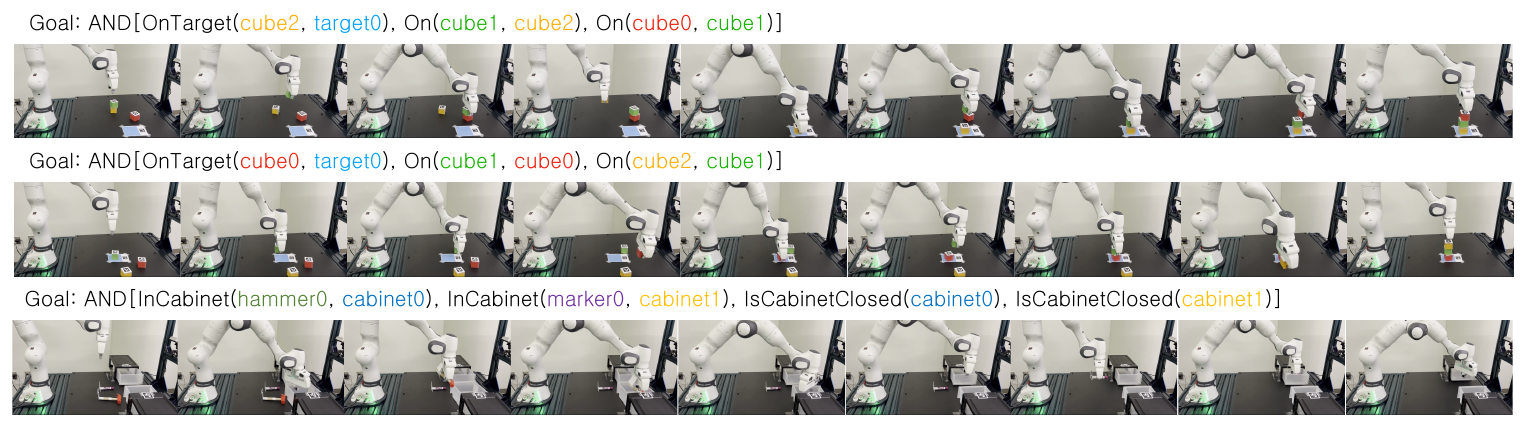}
  \caption{\textbf{Real robot demonstration.} Key frames of three task execution processes (bottom) and their final task goals (top).}
  \label{fig:real_world}
    \vspace{-10pt}
\end{figure*}

\noindent\textbf{Generalize to new task goals.}
Besides evaluating the training goals (shown in Fig.~\ref{fig:sim_setup}), we directly test our models on new task goals for the {StowHammer} and the {PegInHole} domains. For {StowHammer} domain, 
the first test goal is to swap the hammer-cabinet mapping. The second test goal is to place \texttt{hammer1} into \texttt{cabinet0} and keep \texttt{cabinet1} open.
For {PegInHole}, 
the first test goal is to swap the peg-hole mapping. 
The second goal is to only insert \texttt{peg1} into \texttt{hole2}. The results are in Table~\ref{tab:new_task}. We observe that \Ours experiences little performance drop when generalizing to new task goals without additional training, demonstrating strong compositional generalization capability and skill modularity.

\begin{figure}
  \begin{center}
    \includegraphics[width=0.6\linewidth]{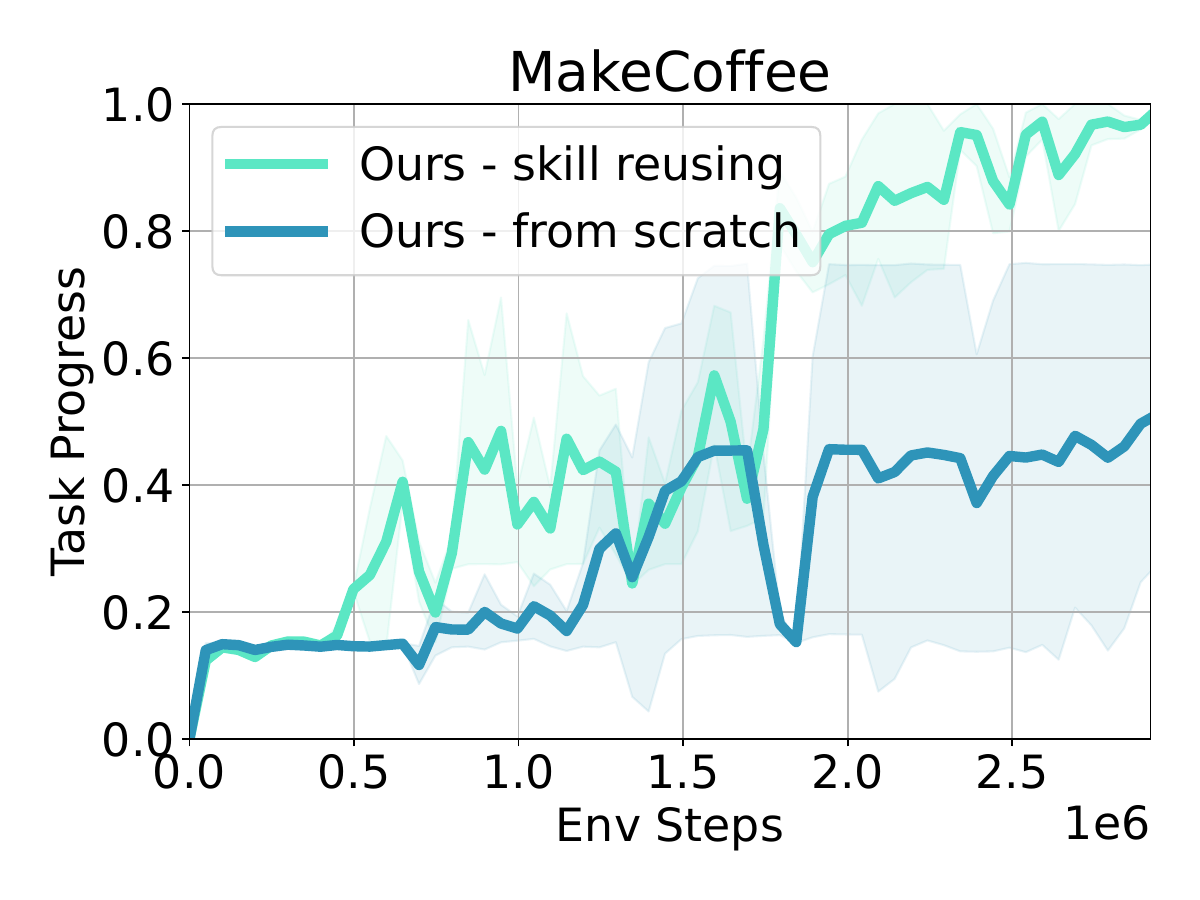}
    \vspace{-10pt}
  \end{center}
  \caption{\textbf{Generalization to new domain.} For {MakeCoffee} task, we compare (a) learning the task from scratch and (b) learning by adapting the skills (\texttt{Pick}, \texttt{Pull}, and \texttt{Push}) learned from the {StowHammer} domain.}
  \label{fig:reusing}
\vspace{-15pt}
\end{figure}

\noindent\textbf{Quick adaptation to new domains.}
Another exciting possibility of \Ours is to transfer skills learned from one domain to another. We design an experiment to validate this feature. The target domain is {MakeCoffee}, which is the most challenging task of the four. We adapt skills \texttt{Pick}, \texttt{Pull}, and \texttt{Push} learned in the {StowHammer} domain for learning the {MakeCoffee} task. As shown in Fig.~\ref{fig:reusing}, compared to learning from scratch, transferring learned skills can significantly accelerate learning (the $x$-axis is shorter than in Fig.~\ref{fig:main_exp}) and enables the robot to solve the entire task. This highlights \Ours's strong potential for continual learning. 


\subsection{Real World Demonstration}
We demonstrate transferring simulation-trained \Ours system to two real-world task domains: \textbf{StackThreeAtTarget} and \textbf{StowObject}. For the {StackThreeAtTarget} task, we randomly place three cubes and a target region on the table. The task is to stack the cubes at the target region. We directly reuse the skills trained in {StackAtTarget} in simulation to demonstrate generalization to different number of objects and initial conditions. The {StowObject} is to stow two objects into two cabins. Similar to {StowHammer}, the task also requires the robot to operate the cabinets. We reuse skills trained in the simulated {StowHammer} domain.

Our system uses a Franka Emika Panda robot. We take RGBD images from an Intel RealSense D435 camera and use AprilTag~\cite{olson2011tags} to detect the 6D poses of task-relevant objects. Our system performs state estimation prior to each skill execution, synchronizes the states to a simulated environment, and executes each skill generated by LEAGUE from the simulated environment through open-loop control.

Fig.~\ref{fig:real_world} shows the key frames of three task execution processes and the corresponding task goals. Our system achieves an 8/10 success rate for the {StackThreeAtTarget} task, and a 6/10 success rate for the {StowObject} task. The failure mode for the {StackThreeAtTarget} task is that the AprilTags getting occluded from the camera in some initial configurations. 
The failure mode for {StowObject} task is that sometimes the learned policy is not able to generate a valid motion for operating the drawer, and the objects slipping from the gripper. 

\section{Conclusions, Limitations, and Future Works}
\label{sec:limit}
We introduced \Ours, an integrated task planning and skill learning framework that represents a virtuous-cycle system: It leverages the high-level reasoning ability and abstraction of a TAMP framework to facilitate the exploration and generalization of an RL skill learner, which in turn expands the capability of the overall system. Through challenging manipulation tasks in both simulation and the real world, we demonstrated that \Ours is effective at solving long-horizon tasks and generalizing to new tasks and domains. 

While empirically effective, our method does have a number of limitations.
As we discussed in Sec.~\ref{ssec:background}, we assume access to a library of skill operators that serve as the basis for skill learning. Relatedly, our assumptions for skill-relevant state abstraction, although effective, may not hold in certain cases (e.g. unintended consequences during exploration). A possible path to address both challenges is to learn skill operators with sparse transition models from unstructured experiences~\cite{silver2023inventing, silver2021learning}. Second, our skill learning process relies on the environment-provided dense reward function. RL algorithms that can better learn from sparse reward would allow \Ours to build a tighter connection with the symbolic space. Finally, in the real-world setting, \Ours is limited by the capability of the off-the-shelf perception algorithms. We plan to explore learning visuomotor control policies to make \Ours easier to deploy in the real world.






\bibliographystyle{plainnat}
{\footnotesize
\bibliography{example}}
\end{document}